\documentclass[10pt,twocolumn,letterpaper]{article}

\usepackage[pagenumbers]{iccv} % To force page numbers, e.g. for an arXiv version

\usepackage{xcolor,colortbl}
\definecolor{tabfirst}{rgb}{1, 0.7, 0.7}
\definecolor{tabsecond}{rgb}{1, 0.85, 0.7}
\definecolor{tabthird}{rgb}{1, 1, 0.7}

\newcommand{\zipnerf}{$\text{Zip-NeRF}^*$}
\newcommand{\diffusionnerf}{$\text{DiffusioNeRF}^*$}
\newcommand{\freenerf}{$\text{FreeNeRF}^*$}
\newcommand{\simplenerf}{$\text{SimpleNeRF}^*$}
\newcommand{\zeronvs}{$\text{ZeroNVS}^*$}
\newcommand{\fsgs}{$\text{FSGS}^\dagger$}
\newcommand{\dngaussian}{$\text{DNGaussian}^\dagger$}
\newcommand{\corgs}{$\text{CoR-GS}^\dagger$}

\newcommand{\reconfusion}{$\text{ReconFusion}^*$}
\newcommand{\duster}{DUSt3R}

\newcommand{\catEd}{$\text{CAT3D}$\xspace}
\newcommand{\maybeappendix}[1]{}

\newcommand\rgt{\aftergroup\mathclose\aftergroup{\aftergroup}\right}

\newcommand{\vect}[1]{{\bf #1}}
\newcommand{\inp}{\vect{I}^{\text{ref}}}
\newcommand{\estinp}{\hat{\vect{I}}^{\text{ref}}}
\newcommand{\estrep}{\hat{\vect{I}}^{\text{nov}}}
\newcommand{\pseudo}{\vect{G}^{\text{nov}}}
\newcommand{\inprep}{\hat{\vect{L}}^{\text{nov}}}

\newcommand{\method}{\textbf{RI3D}}

\definecolor{iccvblue}{rgb}{0.21,0.49,0.74}
\usepackage[pagebackref,breaklinks,colorlinks,allcolors=iccvblue]{hyperref}

\def\paperID{6611} % *** Enter the Paper ID here
\def\confName{ICCV}
\def\confYear{2025}

\title{RI3D: Few-Shot Gaussian Splatting With Repair and Inpainting Diffusion Priors}

\author{
Avinash Paliwal$^{1}$ \hspace{0.9cm}
Xilong Zhou$^{1,3}$ \hspace{0.9cm}
Wei Ye$^{2}$ \hspace{0.9cm}
Jinhui Xiong$^{2}$ \\
Rakesh Ranjan$^{2}$ \hspace{0.9cm}
Nima Khademi Kalantari$^{1}$
\vspace{0.2cm} \\
\normalsize$^1$Texas A\&M University \hspace{0.8cm}
$^2$Meta Reality Labs \hspace{0.8cm}
$^3$ Max Planck Institute for Informatics 
\vspace{0.2cm} \\
}

\begin{document}
% \maketitle
\twocolumn[{
\renewcommand\twocolumn[1][]{#1}
\maketitle
\begin{center}
  \centering
  \vspace{-0.4in}
  \includegraphics[width=\linewidth]{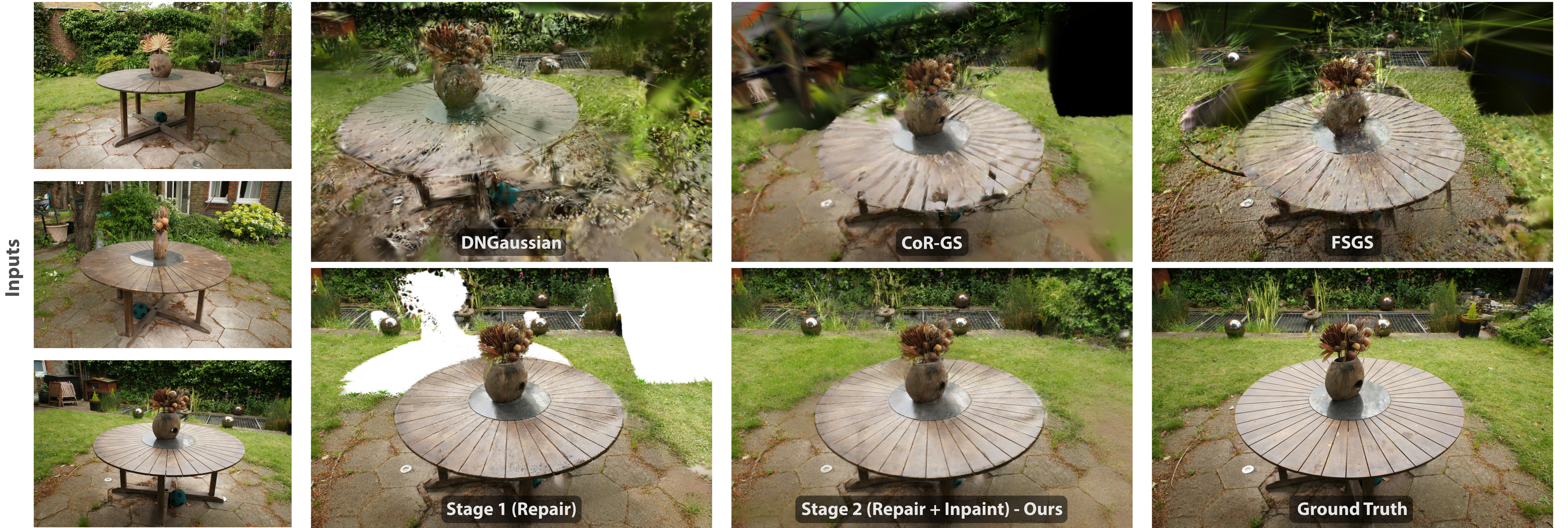}
  \vspace{-0.25in}
  % \vspace{-10pt}
    \captionof{figure}{We introduce a novel sparse view synthesis method that employs two diffusion models, ``repair'' and ``inpainting'', which are responsible for aiding in the reconstruction of visible regions and hallucinating missing regions, respectively. Our approach involves a two-stage optimization process. In the first stage, we use the repair model to constrain the 3DGS optimization and reconstruct the regions covered by the input images. As shown, the output of the first stage properly reconstructs the visible areas, but contains missing regions, which are marked in white. In the second stage, we utilize the inpainting model to fill in these missing areas and continue optimization using the repair model to seamlessly integrate the hallucinated regions with the rest of the scene. Here, we compare our method (``Stage 2'') against several state-of-the-art techniques on a 360\textdegree{} scene using only three input images.}
    \vspace{-0.1in}
  \label{fig:teaser} 
\end{center}
}
]

\begin{abstract}
In this paper, we propose \method{}, a novel 3DGS-based approach that harnesses the power of diffusion models to reconstruct high-quality novel views given a sparse set of input images. Our key contribution is separating the view synthesis process into two tasks of reconstructing visible regions and hallucinating missing regions, and introducing two personalized diffusion models, each tailored to one of these tasks. Specifically, one model ('repair') takes a rendered image as input and predicts the corresponding high-quality image, which in turn is used as a pseudo ground truth image to constrain the optimization. The other model ('inpainting') primarily focuses on hallucinating details in unobserved areas. To integrate these models effectively, we introduce a two-stage optimization strategy: the first stage reconstructs visible areas using the repair model, and the second stage reconstructs missing regions with the inpainting model while ensuring coherence through further optimization. Moreover, we augment the optimization with a novel Gaussian initialization method that obtains per-image depth by combining 3D-consistent and smooth depth with highly detailed relative depth. We demonstrate that by separating the process into two tasks and addressing them with the repair and inpainting models, we produce results with detailed textures in both visible and missing regions that outperform state-of-the-art approaches on a diverse set of scenes with extremely sparse inputs\footnote{{\url{https://people.engr.tamu.edu/nimak/Papers/RI3D}}}.
\end{abstract}    
\section{Introduction}
\label{sec:intro}

The introduction of novel 3D representations, such as neural radiance fields (NeRF)~\cite{mildenhall2021nerf} and 3D Gaussian splatting (3DGS)~\cite{kerbl20233d}, has revolutionized the field of novel view synthesis. While these techniques excel at reconstructing 3D scenes from a large number of images, view synthesis from a sparse set of images remains a challenging problem.

Most state-of-the-art sparse novel view synthesis approaches~\cite{zhu2023fsgs,li2024dngaussian,wang2022sparsenerf} introduce a series of regularizations to constrain the optimization process and avoid overfitting to the input images. While these approaches produce results with reasonable texture in areas visible in a few input images, they often struggle to hallucinate details in the occluded regions. This issue is especially pronounced in 360$^\circ$ scene reconstruction with extremely sparse input images where there are larger missing regions and areas covered by only a single image (see Fig.~\ref{fig:teaser}).

Recently, a couple of techniques~\cite{wu2023reconfusion,gao2024cat3d} propose to address this issue using a diffusion model as a prior during the optimization process. Specifically, they first train a diffusion model~\cite{rombach2022high} on a large multiview image dataset to synthesize novel views given a set of input images. They then perform the optimization using a combination of the input images and novel views, synthesized using this \emph{view synthesis diffusion model}. Although these methods produce significantly better results than the previously discussed techniques, particularly in 360$^\circ$ scenes, they often overblur details, especially in missing areas (see supplementary video). This is because although their view synthesis diffusion models produce visually pleasing novel views, the reconstructed images, particularly in occluded regions, are not 3D-consistent. Using such images during optimization thus leads to overblurring of details. Additionally, since these methods use NeRF as their 3D representation, they suffer from slow rendering times.

To address this issue, we propose to separate the view synthesis process into two tasks of reconstructing the visible and missing areas and introduce two diffusion models, each specialized to help with one of these two tasks. Specifically, inspired by Yang et al.'s approach~\cite{yang2024gaussianobject}, one diffusion model (repair model) is responsible for taking a rendered image as the input and producing a clean image. This model effectively suppresses the artifacts in the reconstructed images during the optimization process. The second diffusion model (inpainting model) is solely responsible for hallucinating details in the missing regions. To ensure these two diffusion models produce results consistent with the scene at hand, we personalize them by tuning the models on the input images.

To ensure fast inference speed, instead of using NeRF as done in the previous approaches~\cite{wu2023reconfusion,gao2024cat3d}, we utilize 3DGS as our 3D representation. To initialize the Gaussians, we propose to obtain per image depth by combining the 3D-consistent, but smooth, depth estimates from DUSt3R~\cite{wang2024dust3r} with the highly detailed relative depth from a monocular depth estimation approach through Poisson blending~\cite{perez2003poisson}. Using the estimated depth, we then assign one Gaussian to each pixel of every input image and project them into 3D space. Moreover, to effectively utilize the two diffusion models, we propose a two-stage optimization process. The goal of the first stage is to reconstruct the scene with detailed texture in the visible areas (see Fig.~\ref{fig:teaser} ``Stage 1'') by utilizing the repair model to enhance the renderings and using them as pseudo ground truth. During the second stage, we hallucinate details in the missing areas using the inpainting model and continue the optimization by utilizing the repair model to seamlessly integrate the hallucinated details with the rest of the scene.

We show that our approach, dubbed \method{}, produces high-quality textures particularly in occluded areas, for challenging scenarios. We further demonstrate that our results outperform the state of the art, both numerically and visually. In summary, we make the following contributions:

\begin{itemize}
    \item To improve the view synthesis process, we propose to utilize two personalized diffusion models: one for enhancing the rendered images and using them as pseudo ground truth during optimization and another for hallucinating details in the missing areas.
    \item We propose a novel approach to initialize Gaussians by combining 3D-consistent, smooth depth maps with highly detailed relative depth from monocular approaches.
    \item We introduce a two-stage optimization strategy that seamlessly incorporates the two diffusion models.
\end{itemize}

\section{Related Work}

\subsection{Radiance Field and 3D Gaussian Splatting}

Neural Radiance Field (NeRF)~\cite{mildenhall2021nerf} is an optimization-based technique for reconstructing 3D scenes from dense input images. The key idea of NeRF is to encode a scene into an implicit neural network that takes a 3D position and view direction as inputs and outputs opacity and view-dependent color. By minimizing the loss between input images and renderings, the implicit neural network is optimized to represent real-world scenes. NeRF has gained considerable attention, inspiring a significant amount of follow-up work to improve rendering quality~\cite{barron2021mip, barron2022mip, barron2023zip, wang2023f2nerf} and efficiency~\cite{muller2022instant, chen2022tensorf, garbin2021fastnerf, reiser2021kilonerf, yu_and_fridovichkeil2021plenoxels}.

The key limitation of NeRF, however, is its slow training and inference, as rendering requires evaluating the network multiple times along a ray. Kerbl et al.~\cite{kerbl20233d} address this problem by proposing 3D Gaussian Splatting (3DGS), which explicitly models the scene using a set of Gaussian primitives. We build our approach on 3DGS, as it produces results comparable to NeRF at a lower computational cost.

\subsection{Sparse Novel View Synthesis}

While both NeRF and 3DGS demonstrate high-quality rendering given dense input image sampling, they struggle with sparse input views. Several techniques have been proposed to address the problem of sparse-input novel view synthesis using NeRF~\cite{deng2022depth, somraj2023simplenerf, wang2023sparsenerf, yang2023freenerf, niemeyer2022regnerf, yu2021pixelnerf, jain2021putting} and 3DGS~\cite{paliwal2024coherentgs, xiong2023sparsegs, zhu2023fsgs} representations. 

More specifically, for NeRF-based methods, PixelNeRF~\cite{yu2021pixelnerf} learns an image-based 3D feature extractor as a prior to optimize NeRF from sparse input views. DietNeRF~\cite{jain2021putting} introduces an auxiliary semantic consistency loss to encourage realistic renderings from novel views. RegNeRF~\cite{niemeyer2022regnerf} proposes geometry and appearance regularization from novel viewpoints. DS-NeRF~\cite{deng2022depth} incorporates depth supervision provided by structure-from-motion into the NeRF pipeline. SparseNeRF~\cite{xiong2023sparsegs} introduces a local depth ranking method and spatial continuity constraints to regularize optimization. FreeNeRF~\cite{yang2023freenerf} regularizes the optimization by reducing the frequency of positional encoding, while SimpleNeRF~\cite{somraj2023simplenerf} provides additional supervision through point augmentation.

Among 3DGS-based methods, FSGS~\cite{zhu2023fsgs} introduces monocular depth supervision and proposes a specially designed densification strategy. SparseGS~\cite{xiong2023sparsegs} presents a novel explicit operator for 3D representations to prune floating Gaussians. CoherentGS~\cite{paliwal2024coherentgs} enhances coherence in the Gaussian representation by constraining the movement of Gaussians and introducing single- and multi-view constraints. DNGaussian~\cite{li2024dngaussian} employs both hard and soft depth regularization to improve sparse-view reconstruction by enforcing surface completeness.

While these methods significantly reduce the number of input images required for high-quality view synthesis, they still struggle to achieve robust results with extremely sparse inputs (e.g., three images). Additionally, they do not provide a reliable approach for reconstructing missing regions and typically fill these areas with overly smooth content.

\subsection{Diffusion-based Methods}

In recent years, Diffusion Models (DM) \cite{ho2020denoising, rombach2022high} have stood out in image generation tasks due to stable training and high-quality results. Because of their success, these models have been used extensively as a prior for various tasks including view synthesis~\cite{lin2024diffbir, shafir2023human, gungor2023adaptive, wang2024exploiting, graikos2022diffusion, Fei2023generative, yang2024gaussianobject, wu2023reconfusion, sargent2023zeronvs, gao2024cat3d}.

The majority of these methods utilize diffusion models for synthesizing novel views of objects. Specifically, GaussianObject~\cite{yang2024gaussianobject} leverages ControlNet~\cite{zhang2023adding} to repair the artifacts introduced by 3DGS optimization and synthesize 3D objects only from four input images. We follow a similar strategy to enhance the quality of rendered images, but focus on reconstructing scenes.

For general scenes, ReconFusion~\cite{wu2023reconfusion} and CAT3D~\cite{gao2024cat3d} train a view synthesis diffusion model, which is further used to regularize NeRF optimization. The main issue with these approaches is that the images reconstructed using the diffusion model are not 3D-consistent, causing the optimized NeRF model to produce blurry results, particularly in the missing areas. In contrast, we introduce two diffusion models to enhance renderings and inpaint missing areas, and we propose a two-stage optimization process that enables the reconstruction of detailed textures in both visible and missing regions.

\section{Background}

In this section, we review two fundamental techniques used in our algorithm: 3D Gaussian splatting and diffusion models. Specifically, we use 3DGS as our 3D representation due to its fast rendering pipeline and diffusion models are used as priors to regularize the optimization process for spare input views.

\begin{figure*}
    \centering
    \includegraphics[width=0.94\linewidth]{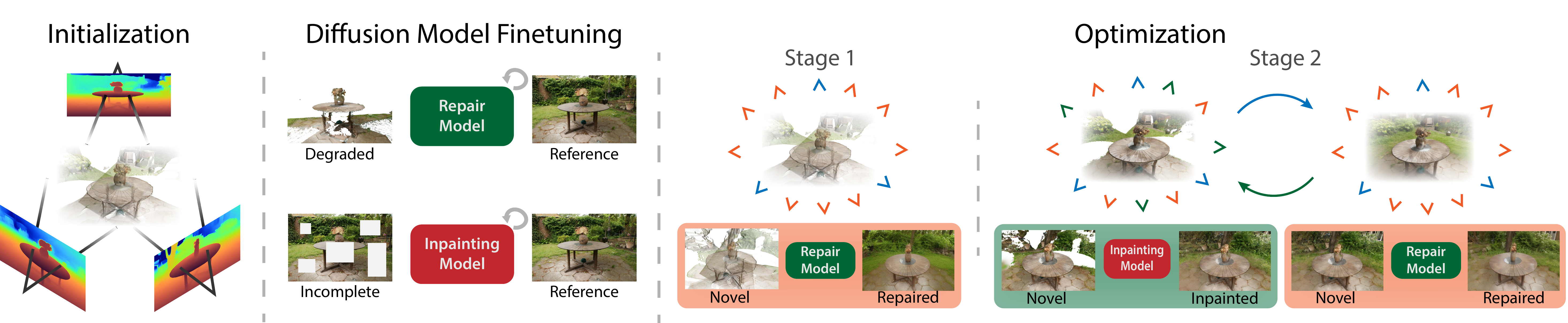}
    \vspace{-0.17in}
    \caption{We provide an overview of the different stages of our approach. First, we initialize the Gaussians by generating high-quality per-view depth maps (Sec.\ref{ssec:init}). Next, we fine-tune the repair and inpainting diffusion models on the scene at hand (Sec.\ref{ssec:personalization}). Finally, we use these models to optimize the 3DGS representation in two stages (Sec.~\ref{ssec:opt}). In the first stage, we reconstruct the areas covered by the input images (blue), using the repair model to generate pseudo ground truth images at $M$ novel views (orange) to constrain the optimization. In the second stage, we first select a subset of novel views (green) to inpaint the missing regions (left) and continue the optimization using the repair model (right). This process of inpainting and optimization is repeated multiple times until all missing areas are reconstructed.}
    \label{fig:overview}
    \vspace{-0.22in}
\end{figure*}

\subsection{3D Gaussian Splatting}

3DGS \cite{kerbl20233d} is a point-based rendering technique representing a scene with a dense set of 3D Gaussians, which can achieve high-quality, fast and differentiable scene rendering. In 3DGS,
each 3D Gaussian is defined using a set of optimizable parameters: 3D position $\vect{x}$, opacity $\sigma$, anisotropic covariance matrix $\Sigma$, and color $\vect{c}$ represented with spherical harmonics (SH) coefficients. Given the representation, point-based $\alpha$-blending is used to render the color $\vect{c}$ at each pixel $\vect{p}$ as follows:

\vspace{-0.15in}
\begin{equation} 
\label{eq:gaussian_render}
	\vect{c}(\vect{p}) = \sum_{i \in N} \vect{c}_{i} \alpha_{i} \prod_{j=1}^{i-1} (1-\alpha_{j}),
\end{equation}
\vspace{-0.15in}

\noindent where $N$ represents the number of Gaussians overlapping with pixel $\vect{p}$, $\vect{c}_{i}$ is view-dependent color computed from SH coefficients of the $i^{\text{th}}$ Gaussian, and $\alpha_{i}$ is the effective opacity, obtained by evaluating the Gaussian and multiplying it with the per-Gaussian opacity $\sigma_{i}$. During optimization, given a set of input images, the parameters of the Gaussian particles are optimized by minimizing the loss between the input and rendered images.

\subsection{Diffusion Models}

Diffusion models~\cite{rombach2022high} are a class of generative models that create data matching the target distribution $q(\vect{x}_0)$ by progressively denoising Gaussian noise $\varepsilon$. Specifically, in the forward process, noise is added to clean data $\vect{x}_{0}$ over $T$ steps, producing a sequence of increasingly noisy data $\vect{x}_0, \dots, \vect{x}_T$. The reverse process then uses the diffusion model to invert this sequence, iteratively denoising from $\vect{x}_T$ back to reconstruct $\vect{x}_0$. Generating high-resolution images with diffusion models is computationally intensive and memory-demanding. To address this issue, latent diffusion models (LDMs)~\cite{rombach2022high} perform the diffusion process in the latent space of a variational autoencoder (VAE)~\cite{kingma2013auto}, reducing memory and computational requirements. Our repair and inpainting models are based on LDMs.

\section{Algorithm}
Given a sparse set of $N$ images $\inp_1, \cdots, \inp_N$ with their corresponding camera poses, our goal is to reconstruct the scene using a 3D Gaussian representation. To do this, we start by initializing the Gaussians using a detailed and 3D-consistent depth estimate at each input image. We then perform a two-stage optimization by utilizing two diffusion models (``repair'' and ``inpainting''), personalized for the scene at hand, to enhance the rendered images and inpaint the missing areas. An overview of our approach is provided in Fig.~\ref{fig:overview}. In the following sections, we discuss our initialization process, the two diffusion models, and the two-stage optimization process. Additional implementation details, such as hyperparameters and learning schedules, can be found in the supplementary materials.

\subsection{3D Gaussian Initialization}
\label{ssec:init}

Initialization is one of the key factors in effectiveness of 3DGS optimization, especially in sparse input settings. Ideally, we would like to start with dense and 3D-consistent Gaussians in areas covered by the input images. A potential solution is to use the point cloud estimated by the multi-view stereo (MVS) technique by Wang et al.~\cite{wang2024dust3r} (\duster{}), which has been shown to perform significantly better than other MVS techniques, particularly with a small number of images. However, \duster's point cloud is usually only accurate in the high-confidence regions where there is overlapping content from multiple input images. For example, for a 360$^\circ$ scene with extremely sparse inputs (e.g., 3 images), the high confidence point cloud generally covers only a small foreground region and is usually sparse. The point cloud in the background areas, which are covered by only a single image, is often highly inaccurate.

To obtain a dense initialization, we instead propose generating per-image depth maps. We then assign a Gaussian to each pixel of every input image and project them into 3D space according to their depth. The key challenge here is obtaining 3D-consistent and detailed depth maps, even in regions covered by only a single image. Our main observation is that the depth estimated by \duster{} (one of the byproducts of this approach) and monocular depth estimation methods, such as Depth Anything V2~\cite{yang2024depth}, have complementary properties (see Fig.~\ref{fig:depth_comp}). Specifically, \duster{} depth in high-confidence regions is 3D-consistent and accurate but smooth, as the estimation is performed on low-resolution images. Additionally, its depth is inaccurate in low-confidence regions. On the other hand, monocular depth is detailed but relative and not 3D-consistent, as the depth maps of each image are obtained independently.

We therefore propose combining the two depth maps to leverage the strengths of both approaches. Specifically, we aim for the combined depth to have two properties: 1) it should preserve the details (edges) from the monocular depth, and 2) it should adhere to the absolute depth values from \duster{} in high-confidence regions. Combining these properties results in the following objective with two terms:

\begin{figure}
    \centering
    \includegraphics[width=\linewidth]{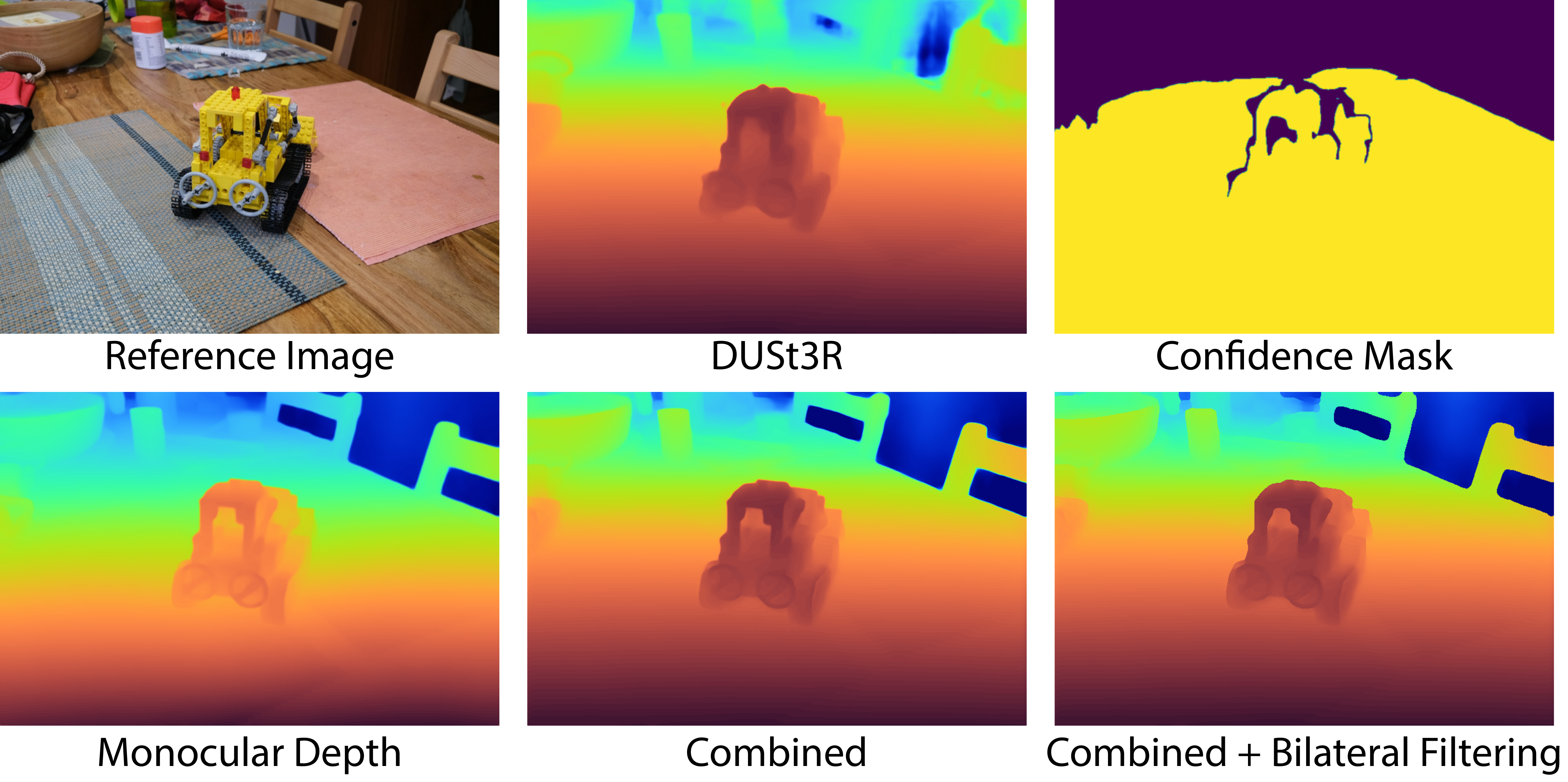}
    \vspace{-0.28in}
    \caption{The depth estimated by \duster{} is geometrically consistent in the high confidence regions (marked in yellow), but of poor quality in the remaining areas. Monocular depth is highly detailed, but is not 3D consistent. Our proposed method combines the two depth maps into a detailed and geometrically consistent depth. Applying bilateral filtering, further sharpens the boundaries.}
    \label{fig:depth_comp}
    \vspace{-0.25in}
\end{figure}

\vspace{-0.1in}
\begin{equation}
\label{eq:blending}
    \vect{d}^* = \arg \min_{\vect{d}} \left[\vect{M} \odot \Vert \vect{d} - \vect{d}^{D} \Vert_2 + \lambda \Vert \nabla \vect{d} - \nabla \vect{d}^{\text{M}} \Vert_2\right],
\end{equation}
\vspace{-0.1in}

\noindent where $\vect{d}^{D}$ and $\vect{d}^{\text{M}}$ are the \duster{} and monocular depth estimates, respectively, and $M$ is a mask indicating the regions where \duster{} has high confidence. The first term enforces similarity between the combined and \duster{} depth maps in high-confidence regions, while the second term ensures that the gradient of the combined depth matches that of the monocular depth everywhere. The weight $\lambda$ controls the contribution of the gradient term, which we set to 10 in our implementation.

Notably, this objective is similar to the Poisson blending equation~\cite{perez2003poisson}, but with a key difference: we enforce the gradient loss everywhere, not just in the low-confidence areas ($1 - M$), since \duster{} depth is smooth, and we aim to incorporate details from monocular depth even in high-confidence regions. As the objective is quadratic, we obtain the solution by solving a linear system of equations (derived by setting the gradient of the objective to zero), which can be efficiently computed using sparse matrix solvers.

Directly using the monocular depth as $\vect{d}^{M}$ in our objective is problematic, as \duster{} and monocular depth can have widely different ranges. To address this, we globally align the monocular depth map to \duster{} depth using a piecewise linear function. Since \duster{} depth is inaccurate in low-confidence regions, we perform the alignment only in high-confidence regions and linearly extrapolate the function for other values. Once we obtain this piecewise linear function, we apply it to the monocular depth to align it with \duster{} depth. We then use this aligned monocular depth as $\vect{d}^{M}$ in Eq.~\ref{eq:blending} to compute the combined depth. Finally, we apply bilateral filtering, as in previous approaches~\cite{Shih3DP20,wang20223d}, to the combined depth to produce per-image depth maps with sharp edges (see Fig.~\ref{fig:depth_comp}).

\begin{figure}
    \centering
    \includegraphics[width=\linewidth]{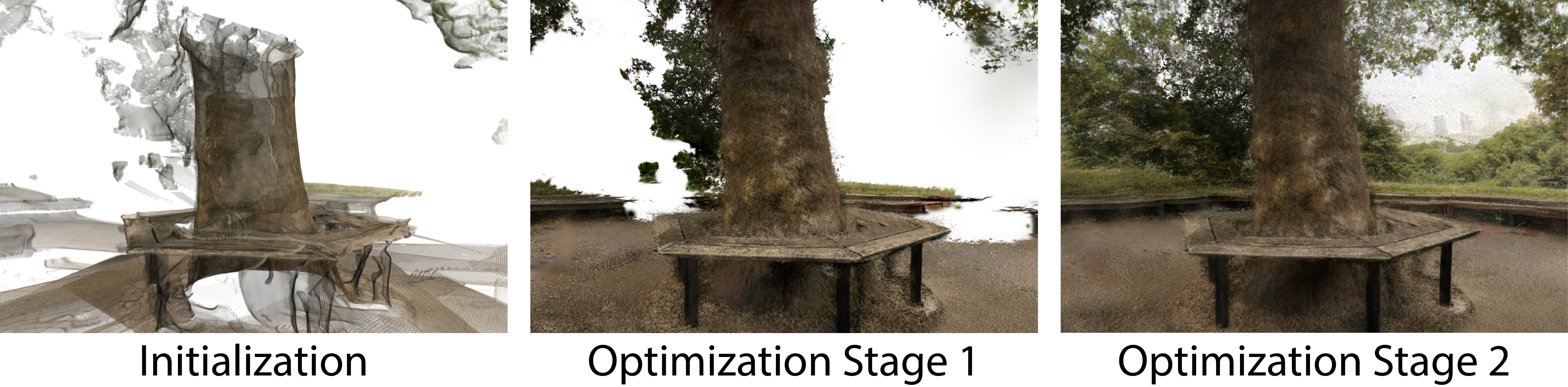}
    \vspace{-0.28in}
    \caption{We show the output at different stages of our approach. Our initialization strategy ensures that Gaussians from different input images are roughly aligned and cover the visible areas of the scene. During the first stage of optimization, we use the repair model to constrain the problem, which in turn helps reconstruct the visible regions with detailed texture. The missing areas are then hallucinated and seamlessly incorporated into the scene during the second and final stage of optimization.}
    \label{fig:stages}
    \vspace{-0.20in}
\end{figure}

Given the per-image depth maps, we assign a Gaussian to each pixel and project it into 3D space along the ray connecting the camera and pixel center~\cite{paliwal2024coherentgs}. We initialize each Gaussian's color using the corresponding pixel color and set the rotation matrix to identity. Additionally, we use an isometric scale, setting it so the projected Gaussians cover 1.4 times the pixel size, and assign an initial opacity of 0.1 to all Gaussians. Figure~\ref{fig:stages} shows our output after initialization.

\subsection{Repair and Inpainting Diffusion Models}
\label{ssec:personalization}

\begin{figure}
    \centering
    \includegraphics[width=\linewidth]{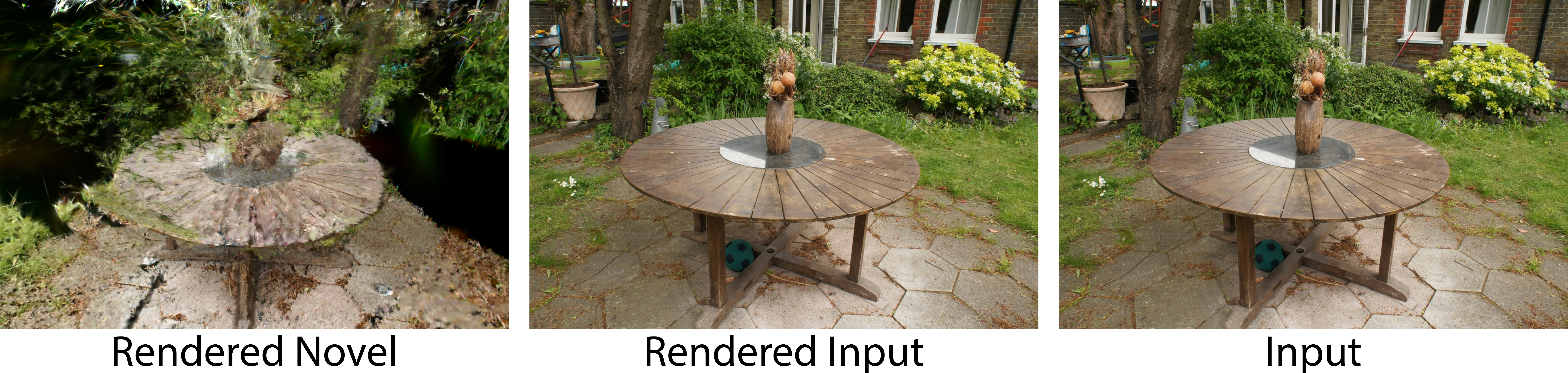}
    \vspace{-0.25in}
    \caption{We show the result of 3DGS optimization using our initialization. In the absence of any constraints, 3DGS optimization quickly overfits to the input images (compare rendered and input images) but produces distracting artifacts in the novel view image. Additionally, unobserved areas will not be reconstructed during optimization, resulting in a dark and blurry appearance. We address these issues using our repair and inpainting models to constrain the optimization and hallucinate missing areas.}
    \label{fig:3dgs_opt}
    \vspace{-0.26in}
\end{figure}

In extremely sparse input settings, 3DGS optimization, even with highly accurate initialization, is brittle and unable to properly reconstruct the scene due to two issues (see Fig.~\ref{fig:3dgs_opt}).
\textbf{First}, the optimization can quickly overfit to the input views, resulting in a representation that produces poor-quality results when rendered from novel camera angles. \textbf{Second}, due to the lack of supervision in occluded regions, these areas are often not reconstructed properly, resulting in blurry or dark appearances in the rendered images.

To address these issues, we propose using two diffusion models. Specifically, the first diffusion model, referred to as the ``repair'' model, takes as input a rendered image from a view not covered by the input cameras and produces a clean version of the image. The key idea is that the repair model can generate clean images for a set of novel views, which can then be used as pseudo ground truth images during optimization to address the first problem. The second diffusion model, referred to as the ``inpainting'' model, tackles the second issue by filling in the missing areas.

Following Yang et al.'s approach~\cite{yang2024gaussianobject}, we use a pre-trained ControlNet~\cite{zhang2023adding} as our base repair model and fine-tune it on the target scene by generating pairs of corrupted and clean images through a leave-one-out strategy. Specifically, we create $N$ subsets of images, each containing $N-1$ images by excluding one input image. We then optimize $N$ separate 3DGS representations on these subsets. After a set number of iterations, we reintroduce the excluded image to its subset and continue optimization for an additional fixed number of iterations. During this process, we render the left-out view at different stages of optimization, pairing these intermediate renderings (corrupted images) with the original left-out image (clean image) to form the training data. Using the corrupted image as the condition for ControlNet and the clean image as ground truth, we train the diffusion model using the standard loss~\cite{yang2024gaussianobject}. Note that by initially excluding an image and later reintroducing it, we generate progressively refined corrupted images, improving the repair model’s ability to handle the final 3DGS optimization (discussed in Sec.~\ref{ssec:opt}).

For inpainting, we use the Stable Diffusion inpainting model~\cite{rombach2022high}. Although this model is powerful, the missing regions in our problem are often large, providing limited context for hallucinating consistent details. To address this, we follow Tang et al.~\cite{realfill}'s approach and fine-tune the diffusion model on the input images by simulating a large set of input-output pairs through random masking.

Since both diffusion models operate on $512 \times 512$ images, we resize the input images during fine-tuning so that their smallest dimension (typically height) is 512 pixels, followed by random $512 \times 512$ cropping to create training data. Resizing is applied first to preserve scene context, which is essential for both the repair and inpainting models, as directly cropping high-resolution images could remove important contextual information. 

\begin{figure*}
    \centering
    \includegraphics[width=0.97\linewidth]{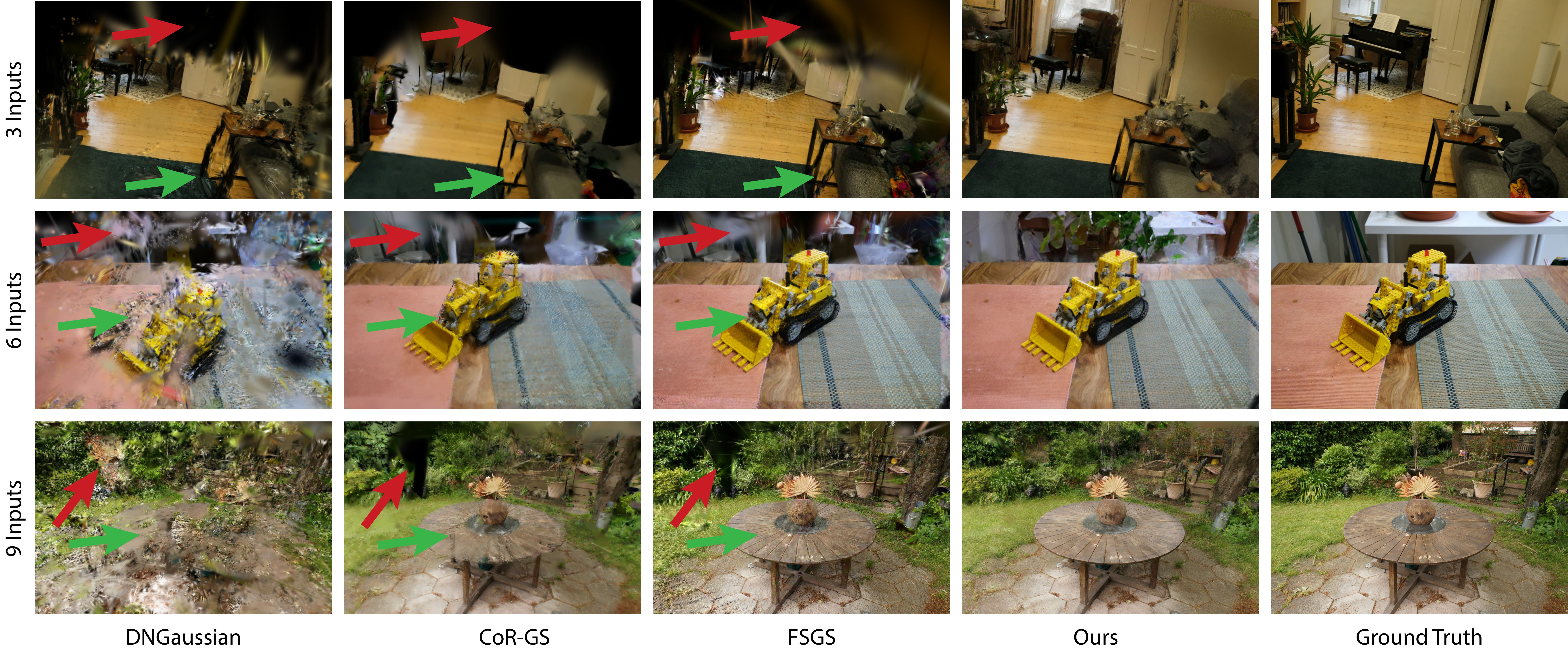}
    \vspace{-0.15in}
    \caption{We compare our approach against the other state-of-the-art methods. Existing methods are not able to properly handle both the visible (green arrows) and missing (red arrows) regions. Through the use of repair and inpainting models, our approach is able to produce high-quality textures in all areas.}
    \label{fig:Comparisons}
    \vspace{-0.25in}
\end{figure*}

\begin{figure}
    \centering
    \includegraphics[width=\linewidth]{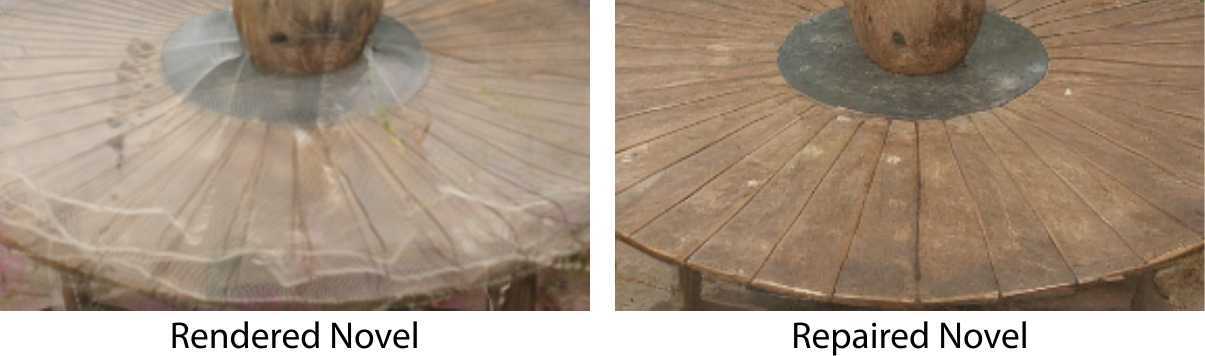}
    \vspace{-0.25in}
    \caption{We show the rendered novel view image of an inset of the Garden scene and the enhanced version using our repaired model. The repaired image is used as the pseudo ground truth during optimization to help with reconstructing detailed textures.}
    \label{fig:garden_repair}
    \vspace{-0.25in}
\end{figure}

\subsection{Optimization}
\label{ssec:opt}

We now have all the elements to begin discussing our proposed optimization strategy, consisting of two stages. The goal of the first stage of the optimization is to obtain a reasonable 3D representation covering only the regions that are visible in the input images. This is a critical step for properly identifying the missing areas. In the second stage, the missing regions are inpainted and the representation is further optimized to obtain a complete scene (see Fig.~\ref{fig:stages}). Below, we discuss the two stages in detail.

\paragraph{Stage 1:} Optimizing solely by minimizing the loss between the rendered and input images can lead to overfitting. To address this, we introduce a set of $M$ novel cameras along an elliptical path aligned with the input cameras. To leverage these novel cameras during optimization, we generate pseudo ground truth images, $\pseudo_j = R(\estrep_j)$, by applying our repair diffusion model $R$ to the rendered images $\estrep_j$. In summary, we optimize the following objective:

\vspace{-0.15in}
\begin{align}
\label{eq:stage1}
    \mathcal{L}_{\text{stage1}} &= \sum_{i = 1}^N\mathcal{L}_{\text{rec}} (\estinp_i, \inp_i) + \sum_{j=1}^M \lambda_j \vect{M}^{\alpha}_j \mathcal{L}_{\text{rec}}(\estrep_j, \pseudo_j) \nonumber \\
    &+ \sum_{j=1}^M \Vert \vect{A}_j \odot (1 - \vect{M}^{\alpha}_j) \odot \vect{M}^b_j\Vert_1.
\end{align}
\vspace{-0.15in}

The first term enforces similarity between the rendered $\estinp_i$ and the input images. The reconstruction loss comprises the original 3DGS losses ($L_1$ and SSIM), as well as LPIPS and depth correlation~\cite{zhu2023fsgs} losses. For the depth correlation loss, we compute the Pearson correlation between the rendered depth from 3DGS and the monocular depth estimated from the input image. The second term encourages similarity between the rendered images and their repaired counterparts from the $M$ novel cameras. The factor $\lambda_j$ represents the camera distance weight~\cite{yang2024gaussianobject}, giving higher weight to novel cameras closer to the input cameras. Additionally, $\vect{M}^{\alpha}_j$ is a binary mask obtained by thresholding the rendered opacity, ensuring that the loss is enforced only in regions covered by the input images. Finally, the third term encourages the rendered opacity $\vect{A}_j$ to be low in the missing regions ($1 - \vect{M}^{\alpha}_j$), as we do not want 3DGS to place any visible Gaussians in these areas. Note that $\vect{M}^b_j$ is a mask identifying the background regions, which we use here because the occluded regions in the foreground are typically small and can be easily inpainted with our repair model. We obtain this background mask by applying agglomerative clustering~\cite{wang20223d} on the monocular depth estimated for repaired images.

To summarize, this objective provides additional supervision using the repair model for $M$ novel cameras to constrain the optimization. Moreover, we focus solely on reconstructing the visible regions by ensuring that the opacity of Gaussians in the missing areas is minimized through the third term. See our output at this stage in Fig.~\ref{fig:stages}.

\paragraph{Stage 2:} At the end of stage 1, the 3DGS representation can reconstruct the visible areas from the input images but lacks information in the missing regions, $(1 - \vect{M}^{\alpha}_j) \odot \vect{M}^b_j$. To address this, we first use our personalized inpainting model to fill in the missing areas in novel camera views and then integrate the hallucinated details into the scene through optimization. Specifically, we select a subset of $K < M$ novel view images with non-overlapping content and inpaint the missing areas using our inpainting diffusion model. We inpaint only a subset of images to prevent independently inpainting overlapping content, which could lead to inconsistent results. We then project the inpainted areas into the scene using monocular depth estimated from the inpainted images. Since monocular depth is relative and has a different range than the reconstructed 3D scene, it cannot be used directly. We address this by combining the monocular depth in the inpainted regions with the rendered depth using Eq.~\ref{eq:blending}.

This process results in $K$ inpainted images, $\inprep_j$. We then apply our repair model to these images, as well as the remaining $M - K$ novel view images, to obtain the pseudo ground truth images, $\pseudo_j$. These images are then used to optimize the 3DGS representation by minimizing the following loss:

\vspace{-0.15in}
\begin{align}
    \mathcal{L}_{\text{stage2}} &= \sum_{i = 1}^N\mathcal{L}_{\text{rec}} (\estinp_i, \inp_i) + \sum_{j=1}^M \lambda_j \mathcal{L}_{\text{rec}}(\estrep_j, \pseudo_j) \nonumber \\
    &+ \sum_{k=1}^K (1 - \vect{M}^{\alpha}_k) \odot \vect{M}^b_k \odot L_p (\estrep_k, \inprep_k).
\end{align}
\vspace{-0.15in}

The first and second terms are similar to those in Eq.~\ref{eq:stage1}. The main difference is that, since most of the scene has now been inpainted, we enforce the loss in the second term across the entire image by removing the visibility mask $\vect{M}^\alpha_j$. The third term ensures that the rendered images $\estrep_j$ remain close to the inpainted images $\inprep_j$ in the inpainted areas, $(1 - \vect{M}^{\alpha}_j) \odot \vect{M}^b_j$. We perform this optimization for a fixed number of iterations, then repeat the inpainting step with a different subset of $K$ images before optimizing again. This iterative process continues until the remaining missing areas in the 3D scene are progressively filled.

At the end of this stage, we obtain a complete 3D representation of the scene, which can be rendered from novel camera views, as shown in Fig.~\ref{fig:stages}.

\section{Results}

We compare our approach against several state-of-the-art methods both visually and numerically and evaluate the effect of various aspects of our proposed method. Here, we show the results on the \textbf{Mip-NeRF 360}~\cite{barron2022mipnerf360} dataset but provide additional results on \textbf{CO3D}~\cite{reizenstein2021common} and video comparisons in the supplementary materials.

\subsection{Visual Results}

In Fig.~\ref{fig:Comparisons}, we compare our approach against several 3DGS-based approaches: DNGaussian\cite{li2024dngaussian}, CoR-GS~\cite{zhang2025cor}, and FSGS~\cite{zhu2023fsgs}. We use the authors' publicly available source code for these methods. As the code for Reconfusion~\cite{wu2023reconfusion} and CAT3D~\cite{gao2024cat3d} is unavailable, we cannot include their results in this figure. However, we show a comparison against their publicly available video results in our supplementary video. As shown in Fig.~\ref{fig:Comparisons}, existing approaches fill the missing areas (red arrows) using either a few large Gaussians or many small Gaussians, leading to a blurry or noisy appearance (the latter in the second and third rows of DNGaussian). In contrast, by utilizing our powerful personalized inpainting model, we are able to reconstruct these areas with detailed textures.

Moreover, existing approaches struggle with properly reconstructing the visible areas (indicated by the green arrows) in these extremely sparse input settings. On the other hand, by providing additional novel view supervision through the repair model, our method is able to reconstruct the visible areas reasonably well. For example, as shown in Fig.~\ref{fig:garden_repair}, our repair model is able to produce high-quality texture on the table from the ghosted rendering. Using these repaired images as pseudo ground truth during optimization enhances texture detail in visible areas.

\begin{table*}[t]
\centering

\caption{We compare our approach against other sparse view synthesis methods. PSNR and SSIM measure pixel-wise error and as such ReconFusion and CAT3D score higher in most cases, since these metrics favor blurrier results. Our LPIPS scores, however, demonstrate that our approach is better at preserving texture details. The numbers for approaches marked by $^*$ are directly grabbed from ReconFusion. CAT3D results are obtained from the original paper. The methods marked by $^\dagger$ are initialized using \duster{} to improve their results. }
\vspace{-0.13in}
\resizebox{0.9\textwidth}{!}{%
\begin{tabular}{clccccccccc}
\cmidrule[\heavyrulewidth]{2-11}
 & 
 & \multicolumn{3}{c}{\textbf{3-view}} & \multicolumn{3}{c}{\textbf{6-view}} & \multicolumn{3}{c}{\textbf{9-view}} \\
 & Method & PSNR $\uparrow$ & SSIM $\uparrow$ & LPIPS $\downarrow$ & PSNR $\uparrow$ & SSIM $\uparrow$ & LPIPS $\downarrow$ & PSNR $\uparrow$ & SSIM $\uparrow$ & LPIPS $\downarrow$ \\ \cmidrule{2-11}

& \diffusionnerf~\cite{Wynn_2023_CVPR} &                      11.05 & 0.189 & 0.735 &                     12.55 & 0.255 & 0.692 &                      13.37 & 0.267 & 0.680 \\                                                                                                                               
& \freenerf~\cite{yang2023freenerf}    &                      12.87 & 0.260 & 0.715 &                     13.35 & 0.283 & 0.717 &                      14.59 & 0.319 & 0.695 \\                                                                                                                               
 & \simplenerf~\cite{somraj2023simplenerf}  &                     13.27 & 0.283 & 0.741 &                     13.67 & 0.312 & 0.721 &                      15.15 & 0.354 & 0.676 \\

 & \zipnerf~\cite{barron2023zip}     & 12.77 & 0.271 &  0.705  & 13.61 & 0.284 & 0.663  & 14.30 & 0.312 & 0.633  \\ 
 & \zeronvs~\cite{sargent2023zeronvs}    & 14.44 & 0.316 & 0.680 & 15.51 & 0.337 & 0.663 & 15.99 &  0.350 &  0.655 \\
 
& \dngaussian~\cite{li2024dngaussian}     & 12.02 & 0.226 &  0.665  & 12.71 & 0.248 & 0.646  & 12.97 & 0.258 & 0.637  \\ 

& \corgs~\cite{zhang2025cor}     & 13.51 & 0.314 &  0.633  & 14.53 & 0.328 & 0.596  & 15.48 & 0.356 & 0.574  \\ 
 & \fsgs~\cite{zhu2023fsgs}     & 13.14 & 0.288 &  \cellcolor{tabthird}0.578  & 14.64 & 0.348 & \cellcolor{tabthird}0.515  & 16.00 & 0.397 & \cellcolor{tabthird}0.470  \\ 
 
 & \reconfusion~\cite{wu2023reconfusion}         &  \cellcolor{tabthird}15.50 &  \cellcolor{tabsecond}0.358 &  0.585  &  \cellcolor{tabsecond}16.93 &  \cellcolor{tabsecond}0.401 &  0.544 &  \cellcolor{tabsecond}18.19 &  \cellcolor{tabthird}0.432 & 0.511    \\
 & \catEd & \cellcolor{tabfirst}16.62 & \cellcolor{tabfirst}0.377 & \cellcolor{tabsecond}0.515 & \cellcolor{tabfirst}17.72 & \cellcolor{tabfirst}0.425 & \cellcolor{tabsecond}0.482 & \cellcolor{tabfirst}18.67 & \cellcolor{tabfirst}0.460 & \cellcolor{tabsecond}0.460 \\
 & \method{} (Ours) & \cellcolor{tabsecond}15.74 & \cellcolor{tabthird}0.342 & \cellcolor{tabfirst}0.505 & \cellcolor{tabthird}16.51 & \cellcolor{tabthird}0.392 & \cellcolor{tabfirst}0.462 & \cellcolor{tabthird}17.48 & \cellcolor{tabsecond}0.439 & \cellcolor{tabfirst}0.415 \\
 \cmidrule[\heavyrulewidth]{2-11}
\end{tabular}
}
\vspace{-0.15in}

\label{tab:main_table}
\end{table*}

\begin{figure*}
    \centering
    \includegraphics[width=\linewidth]{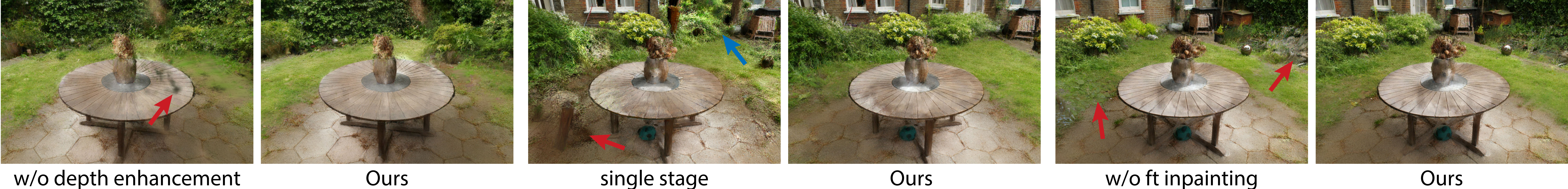}
    \vspace{-0.25in}
    \caption{We show the effect of various components of our system. The differences can be best seen in the supplementary video.}
    \label{fig:ablations}
    \vspace{-0.22in}
\end{figure*}

\subsection{Numerical Results}

In Table~\ref{tab:main_table}, we numerically evaluate our approach by reporting the average error in PSNR, SSIM, and LPIPS~\cite{Zhang2018the} across all scenes in the \textbf{Mip-NeRF 360}~\cite{barron2022mipnerf360} dataset. Our method outperforms all other approaches in PSNR and SSIM, except for ReconFusion and CAT3D, where our results are slightly worse. However, this is primarily because ReconFusion and CAT3D generate smooth, blurry textures in the missing regions (see supplementary video), which are favored by PSNR and SSIM. In contrast, our method achieves the best performance in LPIPS, highlighting its ability to synthesize detailed textures.

\subsection{Ablations}
We evaluate the effect of the core components of our method both visually and numerically in Fig.~\ref{fig:ablations} and Table~\ref{tab:ablations}, respectively. Directly using the \duster{} depth (w/o depth enhancement) results in floating artifacts, as indicated by the red arrow, since \duster{} produces inaccurate depth in low-confidence areas. By combining \duster{} with monocular depth, our approach largely mitigates this issue.

To evaluate the impact of two-stage optimization, we compare our approach against an alternative method where optimization is performed in a single stage using only the repair model. Note that in this case, the repair model is responsible for both enhancing the rendered images and hallucinating details in the missing areas. As seen, the single-stage approach produces floating artifacts due to insufficient constraints to place the inpainted content in the correct 3D location. Additionally, since the repair model is not specifically trained for inpainting, it sometimes hallucinates low-quality textures, as indicated by the blue arrow.

Finally, to assess the impact of personalizing the inpainting diffusion model, we compare our approach against a variant that directly uses our base inpainting model without fine-tuning (w/o ft inpainting). Although the baseline inpainting model generates visually pleasing textures, they sometimes do not perfectly match the surrounding textures. For example, the synthesized textures include a grayish plant and leaf-like texture on the ground (red arrows).

\begin{table}
    \centering
    \caption{We evaluate the effect of different components of our method on the 3-input Mip-NeRF 360~\cite{barron2022mipnerf360} dataset.}
    \vspace{-0.1in}
    \resizebox{0.9\linewidth}{!}{
    \begin{tabular}{lccc}
    \cmidrule[\heavyrulewidth]{1-4}
    & PSNR $\uparrow$ & SSIM $\uparrow$ & LPIPS $\downarrow$ \\
    \cmidrule[\heavyrulewidth]{1-4}
         w/o depth enhancement & 15.61 & 0.336 & 0.513 \\
         single stage & 14.92 & 0.309 & 0.527 \\
         w/o ft inpainting & 15.64 & 0.341 & 0.508 \\
        \method{} (Ours) & \textbf{15.74} & \textbf{0.342} & \textbf{0.505}\\
    \cmidrule[\heavyrulewidth]{1-4}
    \end{tabular}
    }
    % \vspace{-0.1in}

    \vspace{-0.25in}
    \label{tab:ablations}
\end{table}

\section{Limitations}

Our dense initialization, and consequently, the final reconstruction quality, depend on the \duster{} depth. When \duster{} produces a highly inaccurate depth map, our optimization cannot fully correct the misaligned initialization, resulting in ghosting artifacts (see supplementary). However, the modular nature of our approach allows us to replace \duster{} with potentially more accurate models in the future as better methods become available, mitigating this issue. Moreover, unlike approaches like CAT3D~\cite{gao2024cat3d}, our method is not able to handle a single input image, mainly because of the use of leave-one-out strategy for training our repair model. In the future, it would be interesting to avoid this issue by training the repair model on a large scale training data in an offline manner.

\section{Conclusion}

In this paper, we have presented a novel technique based on 3DGS that utilizes diffusion models to reconstruct novel views from a sparse set of input images. Specifically, we initialize 3D Gaussians by proposing a strategy to obtain per-view depth maps by combining multiview stereo depth with monocular depth. We then perform 3DGS optimization using two personalized diffusion models through two stages. We demonstrate that our approach produces better results than the state of the art on challenging scenes.

\section*{Acknowledgements}
The project was funded in part by a generous gift from Meta. Portions of this research were conducted with the advanced computing resources provided by Texas A\&M High Performance Research Computing.
{
    \small
    \bibliographystyle{ieeenat_fullname}
    \bibliography{main}
}
\clearpage
\setcounter{page}{1}
\maketitlesupplementary

\begin{table*}[htbp]
\centering
\caption{We quantitatively compare our approach against other sparse view synthesis methods on CO3D dataset. PSNR and SSIM measure pixel-wise error and as such ReconFusion and CAT3D score higher in most cases, since these metrics favor blurrier results. We outperform the previous approaches in terms of perceptual quality (LPIPS) while being competitive with CAT3D, a concurrent approach. The numbers for approaches marked by $^*$ are directly grabbed from ReconFusion. CAT3D results are also obtained from the original paper. The methods marked by $^\dagger$ are initialized using \duster{} to improve their results. }
\resizebox{0.95\textwidth}{!}{%
\begin{tabular}{clccccccccc}
\cmidrule[\heavyrulewidth]{2-11}
 & 
 & \multicolumn{3}{c}{\textbf{3-view}} & \multicolumn{3}{c}{\textbf{6-view}} & \multicolumn{3}{c}{\textbf{9-view}} \\
 & Method & PSNR $\uparrow$ & SSIM $\uparrow$ & LPIPS $\downarrow$ & PSNR $\uparrow$ & SSIM $\uparrow$ & LPIPS $\downarrow$ & PSNR $\uparrow$ & SSIM $\uparrow$ & LPIPS $\downarrow$ \\ \cmidrule{2-11}

& \diffusionnerf~\cite{Wynn_2023_CVPR} &                         15.65 & 0.575 & 0.597 &                     18.05 & 0.603 & 0.544 &                      19.69 & 0.631 & 0.500 \\
& \freenerf~\cite{yang2023freenerf}    &                         13.28 & 0.461 & 0.634 &                     15.20 & 0.523 & 0.596 &                      17.35 & 0.575 & 0.561 \\
& \simplenerf~\cite{somraj2023simplenerf}  &                     15.40 & 0.553 & 0.612 &                     18.12 & 0.622 & 0.541 &                      20.52 & 0.672 & 0.493 \\                                                                                                                               
  & \zipnerf~\cite{barron2023zip}     & 14.34 & 0.496 & 0.652 & 14.48 & 0.497 & 0.617 & 14.97 & 0.514 & 0.590  \\
 & \zeronvs~\cite{sargent2023zeronvs}    & 17.13 & 0.581 & 0.566 & 19.72 & 0.627 & 0.515 &  20.50 &  0.640 &  0.500 \\

& \dngaussian~\cite{li2024dngaussian}     & 16.95 & 0.497 &  0.463  & 19.59 & 0.601 & 0.425  & 20.68 & 0.647 & 0.408  \\ 

& \corgs~\cite{zhang2025cor}     & 14.09 & 0.487 &  0.501  & 16.93 & 0.545 & 0.463  & 17.86 & 0.544 & 0.458  \\ 
 & \fsgs~\cite{zhu2023fsgs}     & 15.92 & 0.544 &  0.443  & 19.75 & 0.641 & 0.351  & 21.25 & 0.677 & \cellcolor{tabthird}0.316  \\ 

 & ReconFusion~\cite{wu2023reconfusion}         &  \cellcolor{tabsecond}19.59 &  \cellcolor{tabsecond}0.662 &  \cellcolor{tabthird}0.398 &  \cellcolor{tabsecond}21.84 &  \cellcolor{tabsecond}0.714 &  \cellcolor{tabthird}0.342 &  \cellcolor{tabsecond}22.95 &  \cellcolor{tabsecond}0.736 &  0.318 \\
  & \catEd & \cellcolor{tabfirst}20.57 & \cellcolor{tabfirst}0.666 & \cellcolor{tabfirst}0.351 & \cellcolor{tabfirst}22.79 & \cellcolor{tabfirst}0.726 & \cellcolor{tabfirst}0.292 & \cellcolor{tabfirst}23.58 & \cellcolor{tabfirst}0.752 & \cellcolor{tabfirst}0.273 \\
 & \method{} (Ours) & \cellcolor{tabthird}18.72 & \cellcolor{tabthird}0.628 & \cellcolor{tabsecond}0.385 & \cellcolor{tabthird}20.53 & \cellcolor{tabthird}0.673 & \cellcolor{tabsecond}0.311 & \cellcolor{tabthird}21.32 & \cellcolor{tabthird}0.704 & \cellcolor{tabsecond}0.277 \\
 \cmidrule[\heavyrulewidth]{2-11}
\end{tabular}
}

\label{tab:co3d_table}
\end{table*}

\begin{figure}
    \centering
    \includegraphics[width=\linewidth]{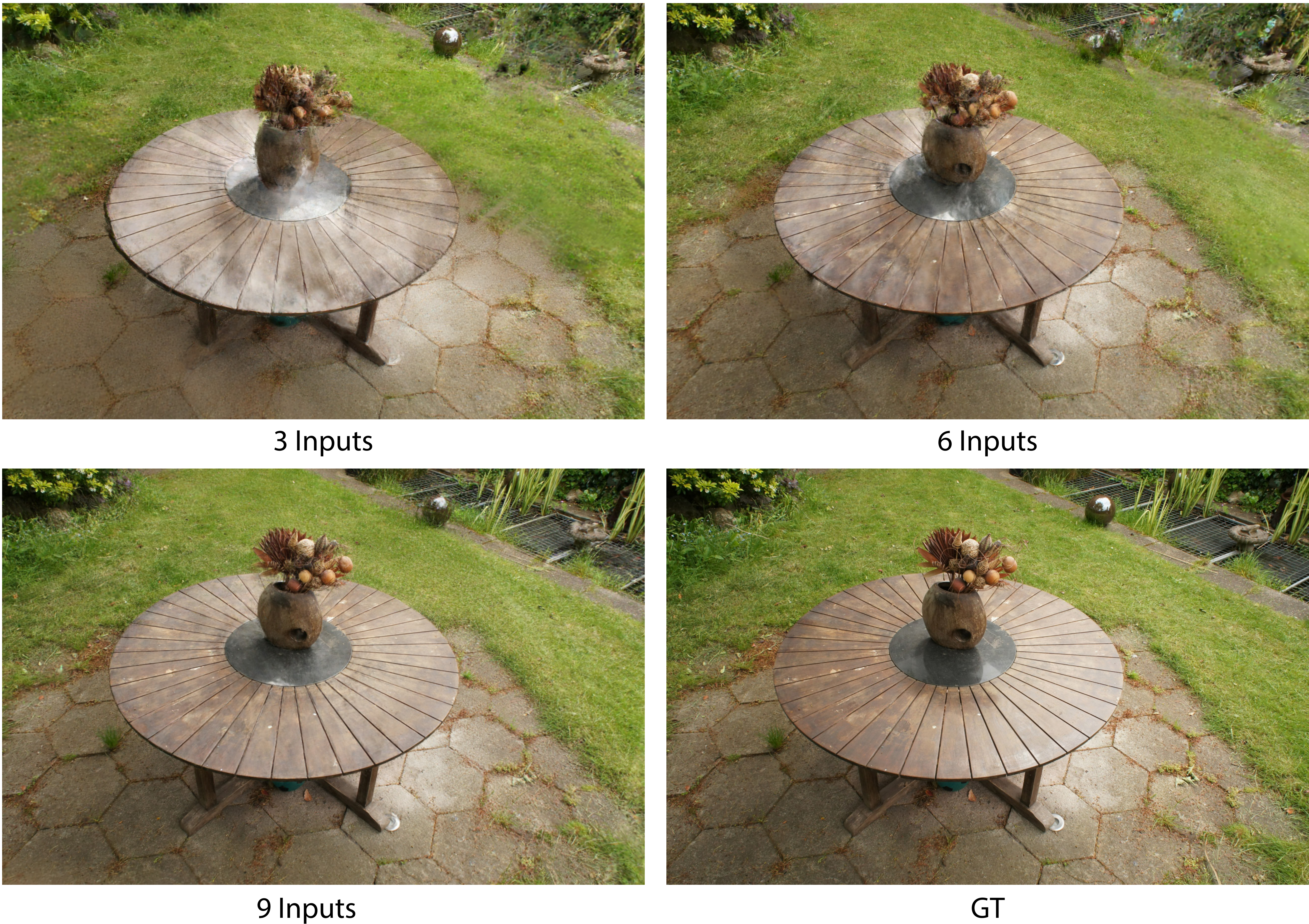}
    \caption{Number of views ablation.}
    \label{fig:num_view_ablation}
\end{figure}

\begin{figure}
    \centering
    \includegraphics[width=\linewidth]{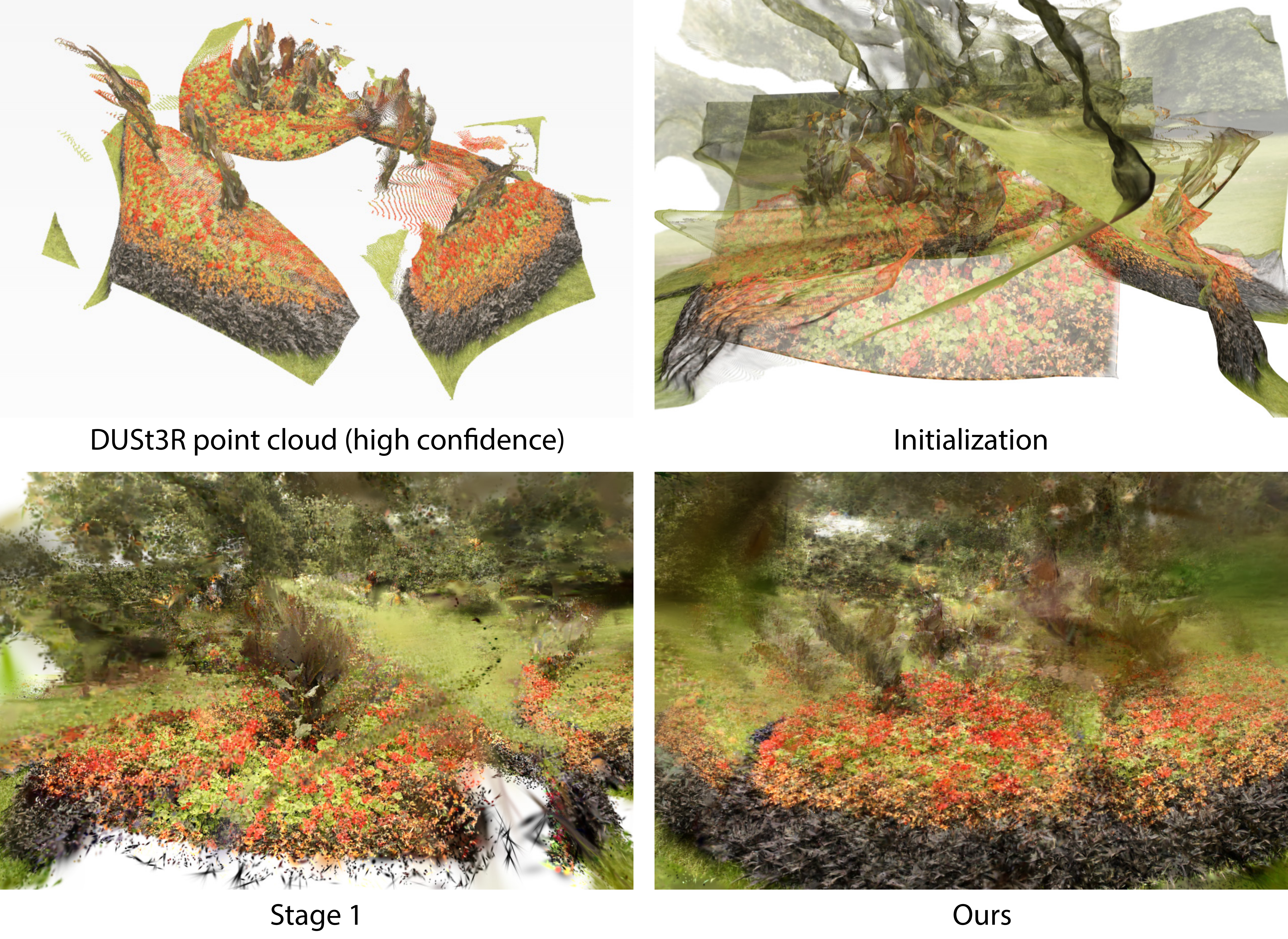}
    \caption{\duster{} sometimes fails to generate a reasonable pointcloud on ambiguous scenes. This in turn affects our optimization quality.}
    \label{fig:limitation}
\end{figure}

\section{Implementation Details}
We provide additional implementation details for some of the models and systems described in the main paper. All optimization and fine-tuning are performed on a single Nvidia RTX A5000 GPU, except for Stage 2, which is carried out using two A5000 GPUs.

\subsection{Repair Model}
Similar to GaussianObject~\cite{yang2024gaussianobject}, the leave-one-out strategy is applied by introducing the left-out view after 6000 iterations of optimization. The optimization continues until iteration 10000, thereby obtaining training pairs for the repair model. We then fine-tune the repair model using these training pairs for 1800 iterations.

\subsection{Inpainting Model}
We adapt the RealFill~\cite{realfill} methodology to fine-tune the Stable Diffusion inpainting model on the sparse input views. Due to unavailability of code by the original authors, we utilize a third-party implementation by Nguyen~\cite{realfill_impl}. Consistent with the referenced GitHub repository, we fine-tune the model for 2000 iterations.

\subsection{Optimization}
We run the \textbf{Stage 1} optimization for 4000 iterations, utilizing 8 evenly distributed novel repaired views in addition to the input training images. The repaired views are refreshed every 400 iterations.

Similarly, \textbf{Stage 2} optimization runs for 4000 iterations. During this stage, we sample 10 evenly distributed views every 200 iterations. For each sampling cycle, we sequentially inpaint and project every other view (5 views) before rendering and repairing all 10 views. Inpainting is performed up to iteration 2800, after which only the repair process is carried out to address minor artifacts.

\section{Additional Results}
\begin{figure*}
    \centering
    \includegraphics[width=\linewidth]{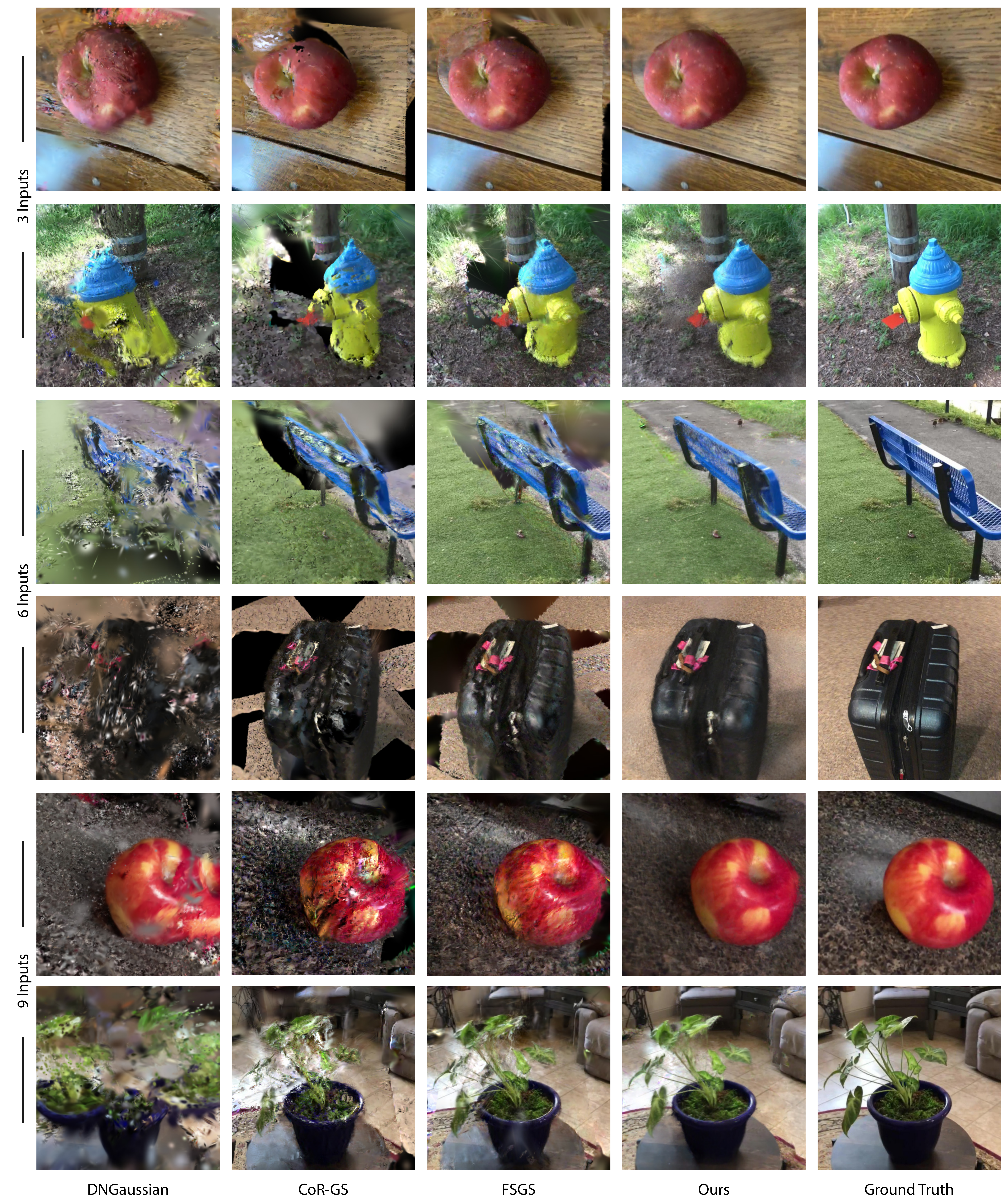}
    \caption{We compare our approach against the other state-of-the-art sparse view synthesis methods on a few scenes from the CO3D dataset.}
    \label{fig:CO3DComparisons}
\end{figure*}

We provide additional results on the \textbf{Mip-NeRF 360}~\cite{barron2022mipnerf360} and \textbf{CO3D}~\cite{reizenstein2021common} dataset for the 3-, 6- and 9-input setting. For both datasets, we utilize the training cameras provided by ReconFusion~\cite{wu2023reconfusion} in our evaluation. In addition to the numerical results presented by ReconFusion and CAT3D~\cite{gao2024cat3d}, we compare against recent state-of-the-art 3D Gaussian based sparse novel view synthesis methods including FSGS~\cite{zhu2023fsgs}, CoR-GS~\cite{zhang2025cor} and DNGaussian~\cite{li2024dngaussian}. In contrast to MipNeRF 360 dataset~\cite{barron2022mipnerf360}, which contains long range general scenes with large missing regions, most CO3D scenes are close up photos of objects in front of a simple background, e.g., a wooden table or floor. This limits the need for and the performance improvement gained from high-quality inpainting. As shown in Table~\ref{tab:co3d_table}, we outperform previous approaches including ReconFusion in terms of perceptual quality (LPIPS) while being competitive with CAT3D. We also provide visual comparisons in Fig.~\ref{fig:CO3DComparisons}. Our approach generates complete and consistent results compared to other 3DGS-based approaches. We provide video comparisons for both MipNeRF 360 and CO3D in the supplementary video.

We provide an additional ablation result (Fig.~\ref{fig:num_view_ablation}) illustrating how the number of training views affects the output quality. As expected, increasing the number of training views leads to better reconstruction quality

\section{Limitations}

As discussed in the main paper, the quality of our dense initialization and consequently the final reconstruction is influenced by the accuracy of the depth maps produced by \duster{}. In cases where \duster{} generates highly inaccurate depth estimates, our optimization procedure cannot fully compensate for the resulting misalignments, leading to noticeable ghosting artifacts, as shown in Fig.~\ref{fig:limitation}. However, due to the modular design of our method, \duster{} can be readily replaced with more accurate depth estimation models as they become available which mitigates this limitation.

\begin{figure*}
    \centering
    \includegraphics[width=\linewidth]{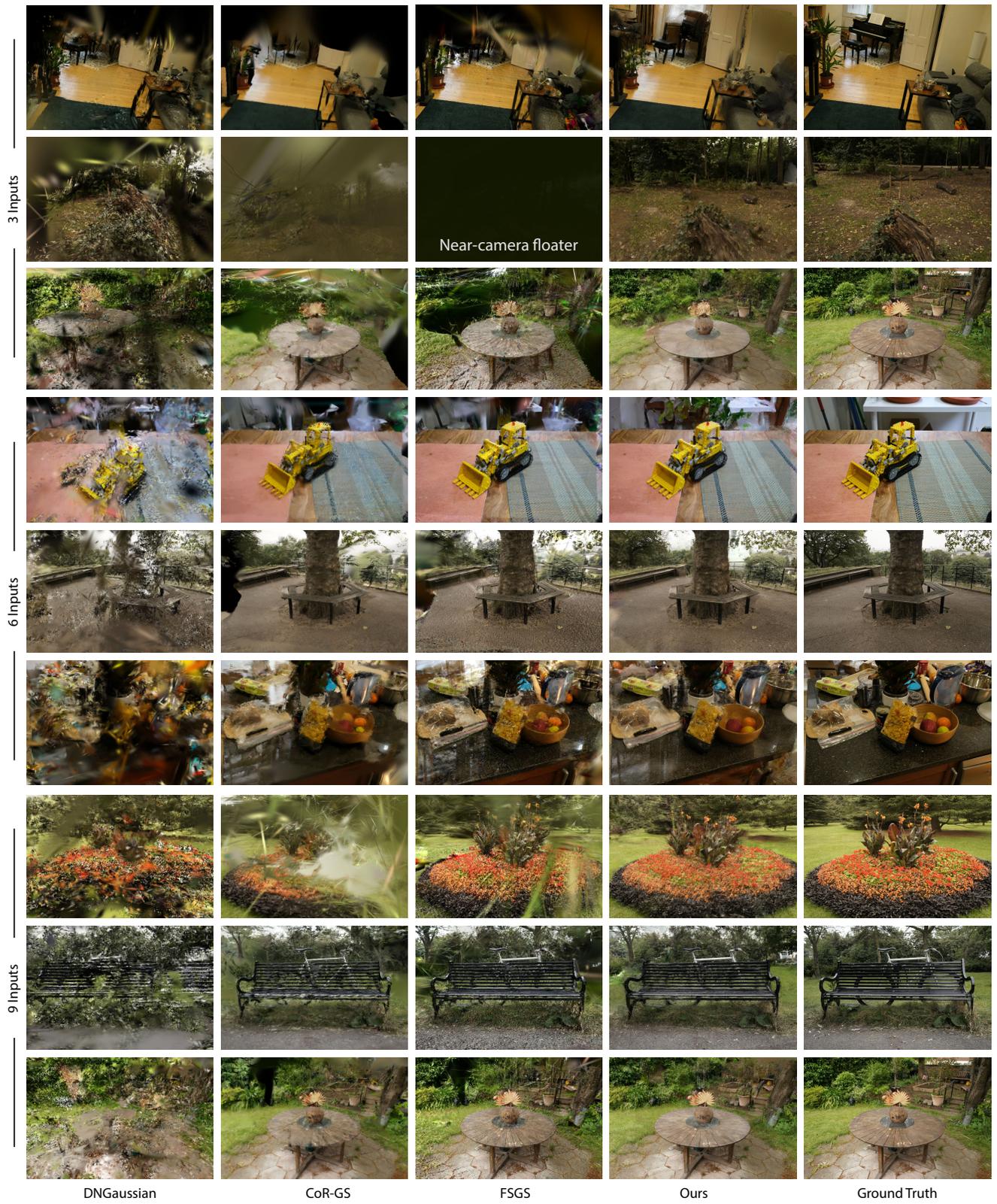}
    \caption{We compare our approach against the other state-of-the-art sparse view synthesis methods on a few scenes from the Mip-NeRF 360 dataset.}
    \label{fig:MIPComparisons}
\end{figure*}

\end{document}